\documentclass[a4paper,review]{cas-sc}

\usepackage[authoryear]{natbib}
\usepackage{graphicx}       
\usepackage{amsmath}
\usepackage{algorithm}
\usepackage{algorithmic}
\usepackage{wrapfig}
\usepackage{multirow}
\usepackage{array}
\usepackage{threeparttable}

\usepackage{lineno} 

\def\tsc#1{\csdef{#1}{\textsc{\lowercase{#1}}\xspace}}
\tsc{WGM}
\tsc{QE}
\tsc{EP}
\tsc{PMS}
\tsc{BEC}
\tsc{DE}

\begin{document}


\let\WriteBookmarks\relax
\def\floatpagepagefraction{1}
\def\textpagefraction{.001}
\shorttitle{Learning to search for VRPMTW}

\title [mode = title]{Learning to Search for Vehicle Routing with Multiple Time Windows}         








\begin{abstract}
In this study, we propose a reinforcement learning-based adaptive variable neighborhood search (RL-AVNS) method designed for effectively solving the Vehicle Routing Problem with Multiple Time Windows (VRPMTW). Unlike traditional adaptive approaches that rely solely on historical operator performance, our method integrates a reinforcement learning framework to dynamically select neighborhood operators based on real-time solution states and learned experience. We introduce a fitness metric that quantifies customers' temporal flexibility to improve the shaking phase, and employ a transformer-based neural policy network to intelligently guide operator selection during the local search. Extensive computational experiments are conducted on realistic scenarios derived from the replenishment of unmanned vending machines, characterized by multiple clustered replenishment windows. Results demonstrate that RL-AVNS significantly outperforms traditional variable neighborhood search (VNS), adaptive VNS (AVNS), and state-of-the-art learning-based heuristics, achieving substantial improvements in solution quality and computational efficiency across various instance scales and time window complexities. Particularly notable is the algorithm's capability to generalize effectively to problem instances not encountered during training, underscoring its practical utility for complex logistics scenarios.

\end{abstract}


\thispagestyle{empty} 

\begin{center}
{\Large \bf{Learning to Search for Vehicle Routing with Multiple Time Windows}}

\vskip 0.6cm
{\bf Kuan Xu$^{a}$, Zhiguang Cao$^{b}$, Chenlong Zheng$^{a}$, Lindong Liu$^{a}$\footnote{Corresponding author. ldliu@ustc.edu.cn}} \\ 
\vskip 0.6cm

{\small $^a$International Institute of Finance, School of Management, \\University of Science and Technology of China, 230026, P.R. China}\\
\vskip 0.4cm
{\small $^b$School of Computing and Information Systems, \\Singapore Management University, 178902, Singapore}\\

\end{center}

\vskip 2cm
\textbf{Acknowledgements:}
The work was supported by the National Natural Science Foundation of China [Grants 72471216, 72022018, 72091210] and Youth Innovation Promotion Association, Chinese Academy of Sciences [Grant No. 2021454].

\newpage
\begin{highlights}
\item A novel RL-AVNS approach integrates reinforcement learning with adaptive variable neighborhood search for solving VRPMTW.
\item A specialized fitness metric quantifying customers' temporal flexibility enhances the shaking phase effectiveness.
\item Computational experiments on realistic unmanned vending machine replenishment scenarios demonstrate RL-AVNS's superior performance.
\item The approach exhibits strong generalization capabilities to unseen problem instances, offering practical value for complex logistics optimization.
\end{highlights}

\begin{keywords}
Vehicle routing \sep Multiple time windows \sep Reinforcement learning \sep Unmanned vending machine replenishment
\end{keywords}

\maketitle

\section{Introduction}

Vehicle Routing Problems (VRPs) are fundamental to optimizing logistics and transportation systems. They are critical for ensuring timely and cost-effective deliveries in various industries, including e-commerce, healthcare, and food services \citep{toth2014vehicle,cordeau2000vrptw}. In response to growing customer expectations for personalized services, logistics providers are increasingly offering flexible delivery options to improve service quality and maintain a competitive edge. This is especially true in industries like e-commerce and retail. In these sectors, companies provide multiple delivery time windows for customers to choose their preferred slots. Such evolving demands have necessitated the development of advanced VRP variants, particularly the Vehicle Routing Problem with Multiple Time Windows (VRPMTW). These variants allow logistics systems to accommodate both customer preferences and operational requirements \citep{beheshti2015vehicle}. 

Additionally, VRPMTW has significant applications in long-haul transportation. In this scenario, strict safety regulations, such as mandatory rest periods for truck drivers, further complicate the scheduling and routing process \citep{rancourt2013long}. The flexibility offered by VRPMTW is crucial for managing complex customer demands in real-world scenarios. This enables logistics systems to simultaneously address customer preferences and regulatory compliance, and thereby enhances customer satisfaction and loyalty while reducing operational costs through optimizing resource allocation.

Exact algorithms for solving VRPs, such as branch-and-cut and branch-and-price \citep{vrpsolver}, ensure optimality for small instances. But they are often impractical for larger problems due to their exponential computational demands. In contrast, heuristic methods include construction heuristics like the saving algorithm \citep{clarke1964saving} and improvement heuristics such as 2-opt* \citep{potvin1995optstar}. These methods provide more efficient solutions by trading optimality for scalability. Local search is a prevalent improvement heuristic that iteratively refines routes by exploring neighboring configurations. It efficiently produces high-quality solutions for large-scale VRPs \citep{groer2010library}. Some of the most widely recognized approximate approaches include genetic algorithms and Variable Neighborhood Search (VNS).

VNS further improves upon local search by systematically changing neighborhood structures. It incorporates a shaking mechanism to escape local optima and explore larger solution spaces more effectively, making it well-suited for addressing VRPs with complex constraints \citep{gendreau2019handbook}. For VRPMTW specifically, VNS effectively handles multiple time windows by thoroughly exploring the solution landscape \citep{braysy2003reactive,stenger2013adaptive,wei2015variable}. Successful implementations often employ adaptive neighborhood selection based on performance history, allowing the algorithm to favor more promising search directions \citep{karakostas2020adaptivea,bezerra2023variable}. However, manual configuration often leads to inefficiencies, which highlights the need for an adaptive approach to enhance the effectiveness of VNS.

On the other hand, in the field of artificial intelligence, deep Reinforcement Learning (RL) has been successfully applied to a wide range of decision-making problems. They include playing complex board games like AlphaGo \citep{silver2017mastering}, predicting protein structures in AlphaFold \citep{jumper2021highly} and enabling advancements in autonomous driving \citep{kiran2021deep}. In recent years, deep RL has been successfully applied to solve combinatorial optimization problems and demonstrate its capability to learn optimal strategies in complex environments. It has been used to directly predict solutions end-to-end, as seen in neural combinatorial solvers for routing problems \citep{bello2017neural}. It also enhances traditional operations research algorithms, making them more adaptive and efficient \citep{morabit2021machine}. Configuring neighborhood operators in VNS is challenging, and RL offers a promising solution by dynamically adapting the ordering of these operators according to the problem context. This adaptability addresses the inefficiencies associated with manual configuration and allows for more effective and efficient exploration of the solution space. Despite its considerable potential, to the best of our knowledge, no prior attempts have been made to apply deep RL to solve VRPMTW.

To effectively tackle the complex constraints inherent in VRPMTW, we integrate traditional search-based heuristics with deep neural networks. Specifically, we propose RL-AVNS, a hybrid method that uses an RL agent to adaptively control the iterative process of VNS. In this approach, we employ a policy network to dynamically select operators in VNS according to the current solution and historical performance. The neighborhood operators in RL-AVNS include one type of perturbation operator to diversify the search by escaping local optima, and multiple types of improvement operators to intensify the search and refine solutions. To effectively manage the multiple time window constraints, we have also introduced a new evaluation mechanism. This helps improve the effectiveness of the improvement operators, better handle the complexities of multiple time windows and ultimately enhance solution quality. Our method integrates the learning capabilities of deep neural networks with the strategic search potential of VNS. It addresses the limitations of traditional methods and effectively adapts to the evolving solution context of VRPMTW. Extensive experiments in single and multiple time window scenarios demonstrate that RL-AVNS outperforms existing learning-based neural heuristics and traditional non-learning heuristics.

Our contributions are summarized as follows.
\begin{itemize}
    \item We propose RL-AVNS, which is the first attempt of deep reinforcement learning to solve VRPMTW, providing an adaptive neighborhood selection mechanism for variable neighborhood search algorithm.
    \item We develop a new evaluation metric specifically designed for multiple time window constraints, which improves the effectiveness of the shaking operator and enhances constraint management.
    \item We conduct comprehensive experiments demonstrating that our proposed RL-AVNS outperforms existing approaches, highlighting its potential for real-world logistics optimization.
\end{itemize}

\section{Literature review}

\subsection{VRPMTW: applications and approaches}
The VRPMTW extends the classic VRP to address real-world delivery operations where customers provide multiple service windows. This variant appears in diverse applications including industrial gas distribution \citep{pesant1999flexibility}, long-haul transport with driver rest periods \citep{goel2012minimum, rancourt2013long}, and furniture and electronic goods delivery requiring real-time route adjustments \citep{belhaiza2019three}.


In these contexts, the primary objectives include minimizing total cost while serving customers within their specified time windows and respecting vehicle capacity constraints. Several studies also focus on minimizing route duration to improve fuel efficiency and operational costs. \citet{ferreira2018} highlighted the importance of this approach for tightly scheduled logistics operations, while \citet{larsen2019fast} demonstrated its applicability to large distribution networks where small inefficiencies can compound into significant operational losses.

The complexity of VRPMTW lies in the discontinuities introduced by multiple time windows, which significantly complicate the optimization process. This has led to diverse solution methodologies evolving over time. Early approaches include the ant colony system by \citet{favaretto2007ant}, which demonstrated the potential of bio-inspired algorithms for this problem. A significant development came from \citet{belhaiza2014hybrid}, who introduced a hybrid variable neighborhood-tabu search heuristic that integrated adaptive memory strategies to enhance solution quality.


Recent research has expanded these approaches in several directions. \citet{beheshti2015vehicle} addressed multi-objective optimization, balancing travel costs and customer satisfaction through genetic algorithms. \citet{larsen2019fast} improved computational efficiency with fast delta evaluation methods for adaptive large neighborhood search. \citet{hoogeboom2020efficient} contributed exact polynomial algorithms for determining optimal service start times within multiple time windows, effectively minimizing route durations. Similarly, \citet{schaap2022large} employed large neighborhood search techniques combined with dynamic programming to optimally select service windows, effectively reducing travel distances and durations. \citet{bogue2022column} introduced a hybrid approach using column generation with VNS to establish robust lower bounds and improve solutions through post-optimization.

Table~\ref{tab:review1} provided a summary of the related literature alongside our study.

\begin{table}[h!]
\centering
\caption{Comparison between the relevant literature and the current study}
\label{tab:review1}
\renewcommand{\arraystretch}{1.25}
\scalebox{0.87}{
\begin{tabular}{@{}p{3cm}>{\raggedright\arraybackslash}p{3cm}p{2cm}p{9cm}@{}}
\toprule
\textbf{References}             & \textbf{Objective}                    & \textbf{Algorithm}      & \textbf{Main case scenario} \\ \midrule
\citet{favaretto2007ant}        & Minimum-duration                      & ACO           & Synthetic instances based on \citet{fisher1994optimal}    \\
\citet{belhaiza2014hybrid}      & Minimum-cost or minimum-duration      & HVNTS         & Synthetic instances based on \citet{solomon1987algorithms}    \\
\citet{beheshti2015vehicle}     & Minimum-cost and maximum satisfaction & CCMQGA        & A distribution company optimizes juice delivery routes with time windows, customer satisfaction, and minimal travel cost using homogeneous vehicles.    \\
\citet{ferreira2018}            & Minimum-duration                      & VNS           & Synthetic instances from \citet{belhaiza2014hybrid}    \\
\citet{larsen2019fast}          & Minimum-cost or minimum-duration      & ALNS + DP     & Synthetic instances from \citet{belhaiza2014hybrid}    \\
\citet{belhaiza2019three}       & Minimum-cost                          & MSDEH         & A real-life cash in transit problem provided by a Canadian transportation and logistics research oriented company.    \\
\citet{schaap2022large}         & Minimum-cost or minimum-duration      & LNS + DP      & The new designed large-scale instances that reflect planing tasks in user-centered last-mile logistics.    \\
\citet{bogue2022column}         & Minimum-cost                          & CG + VNS      & Synthetic instances based on \citet{solomon1987algorithms}    \\
This study                      & Minimum-cost and minimum-duration     & RL + AVNS     & Unmanned vending machine replenishment with multiple time windows based on real-world operational constraints and distribution patterns.    \\ 
\bottomrule
\end{tabular}
}
\end{table}

\subsection{Learning heuristics: constructive and iterative solvers}

In recent years, RL has emerged as a promising approach for combinatorial optimization problems. Unlike traditional methods, RL learns effective decision-making policies directly from data, adapting dynamically to complex environments \citep{survey2021ml4co}. This capability has led to successful applications in routing, scheduling, and resource allocation \citep{survey2021rl4co, survey2022ml4metaheu}, particularly in intelligent transportation systems where efficient routing and resource allocation are crucial \citep{survey2021rl4transportation, survey2024ml4vrp}. The versatility and adaptability of RL make it an ideal tool for addressing the complexities of intelligent transportation, ultimately improving efficiency, reducing congestion, and enhancing the overall quality of urban mobility solutions.

RL neural heuristics for VRPs can be categorized into constructive and iterative types based on their solution generation approaches. Neural constructive solvers sequentially generate solutions by incrementally adding nodes, efficiently constructing complete routes \citep{nazari2018reinforcement, xin2021multi, kwon2020pomo}. For VRP with (single) time window (VRPTW), the Joint Attention Model for Parallel Route-Construction (JAMPR) \citep{JAMPR} extended \citet{kool2018attention}'s attention model by pioneering constructive neural heuristics with advanced masking mechanisms to handle time window constraints. \citet{MUSLA} further explored multi-step decision-making to enhance solution feasibility for TSP with time windows, though challenges remained in effectively addressing complex constraints. While constructive methods offer simplicity and speed, they struggle with complex constraints due to limited contextual information, typically relying on masking mechanisms and penalties to handle time window constraints.

In contrast to constructive approaches, neural iterative solvers address VRPs by iteratively refining solutions with search operators \citep{lu2019learning, xin2021neurolkh, wu2022learning, ma2024learning}. For time-constrained variants, \citet{SoftLR} employ Lagrange relaxation to penalize infeasibility, while \citet{gao2020learn} propose frameworks for designing local-search heuristics. \citet{chen2020dynamic} present a neural heuristic called dynamic partial removal, using hierarchical recurrent graph convolutional networks to manage destruction and repair phases. Recently, \citet{zong2024reinforcement} augmented rewards with time-window penalties to progressively learn and adhere to constraints during training.


RL techniques have been extended to various VRP variants with time constraints. \citet{zhang2020multiagent} present a multi-agent framework for VRP with soft time window constraints. Addressing sustainable transportation, \citet{lin2022deep} extend these methodologies to electric vehicle routing, balancing battery charging with time-sensitive deliveries. \citet{nazari2018reinforcement} introduce an architecture for generic VRP instances, adapting reward structures to enforce both soft and hard constraints.


For dynamic routing, \citet{hildebrandt2023opportunities} explore RL for real-time decision-making under stochastic conditions with hard constraints. \citet{mozhdehi2024edge} propose deep RL solutions for heterogeneous electric vehicle routing with time windows using temporal embeddings. Recognizing complementary strengths, \citet{mao2020hybrid} combine deep RL with local search heuristics to address complex constraints.


Collectively, these studies demonstrate the growing efficacy of RL in addressing VRPs challenges. Key benefits included the ability to dynamically adapt to changing environments, to effectively handle hard constraints like time windows, and to integrate additional considerations such as energy consumption and stochastic conditions. These advancements paved the way for innovative and practical routing solutions that were both efficient and applicable to real-world scenarios.

\section{Mathematical model formulation}
\label{sec:model}

\begin{table}[]
    \renewcommand{\arraystretch}{1.25}
    \centering
    \caption{Notations}
    \label{tab:notations}
    \scalebox{0.99}{
    \begin{tabular}{c|l}
    \hline
    \textbf{Symbol} & \textbf{Description} \\
    \hline
    $V$ & Set of all nodes, including the depot node (indexed by 0). \\
    $C$ & Set of customers excluding the depot, $C = V \setminus \{0\}$. \\
    $K$ & Set of vehicles available for routing. \\
    $W_i$ & Set of time windows for customer $i$, where $m\in W_i$ indexes the time windows defined by $[e_{im}, l_{im}]$. \\
    $Q$ & Maximum capacity of each vehicle. \\
    $M$ & a sufficiently large positive constant. \\
    $c_{ij}$ & Travel length (time) from customer $i$ to customer $j$. \\
    $s_i$ & Service time required at customer $i$. \\
    $d_i$ & Demand of customer $i$ that must be satisfied. \\
    $x_{ijk}$ & Binary variable that equals 1 if vehicle $k$ travels directly from customer $i$ to customer $j$, 0 otherwise. \\
    $y_{im}$ & Binary variable that equals 1 if customer $i$ is served during time window $m$, 0 otherwise. \\
    $a_i$ & Arrival time at customer $i$. \\
    $t_i$ & Time when the service begins at customer $i$. \\
    \hline
    \end{tabular}
    }
    \label{tab:notations}
\end{table}

The VRPMTW expands upon the VRPTW framework, which requires serving each customer $i$ within a specific time window $[e_i, l_i]$. VRPTW introduces timing constraints alongside vehicle capacity limitations, significantly increasing routing complexity. The goal remains minimizing total route cost while ensuring timely service within specified windows without exceeding vehicle capacities. This complexity necessitates careful sequencing of customer visits to satisfy both capacity constraints and time requirements.

VRPMTW extends this model by allowing each customer to have multiple acceptable time windows, adding greater complexity and realism. In VRPMTW, each customer may be available for service during several distinct time windows, which better reflects real-world scenarios, such as businesses with different operating hours or residents with varying availability throughout the day. This inclusion of multiple time windows significantly heightens the computational challenge, requiring more sophisticated routing algorithms to determine optimal routes. Based on the notation in Table \ref{tab:notations}, we present the mathematical formulation of VRPMTW as follows:

\begin{align}
\text{Minimize} \quad 
& Z = \sum_{k \in K} \sum_{i \in V} \sum_{j \in V} c_{ij} x_{ijk} \label{eq:obj} \\
\text{subject to} \quad 
& \sum_{k \in K} \sum_{j \in V} x_{ijk} = 1,                            && \forall i \in C, \label{eq:customer_visit_once} \\
&\sum_{j \in V} x_{0jk} = 1,                                            && \forall k \in K, \label{eq:strat_at_depot} \\
&\sum_{i \in V} x_{i0k} = 1,                                            && \forall k \in K, \label{eq:end_to_depot} \\
& \sum_{j \in V} x_{hjk} = \sum_{i \in V} x_{ihk},                      && \forall h \in C, k \in K, \label{eq:flow_conservation} \\
& \sum_{i \in C} d_i \sum_{j \in V} x_{ijk} \leq Q,                     && \forall k \in K,  \label{eq:capacity} \\
& t_{i} + s_{i} + c_{ij} - M(1 - \sum_{k \in K} x_{ijk}) \leq t_{j},    && \forall i, j \in V,  \label{eq:subtour_elimination} \\
& \sum_{m \in W_i} y_{im} = 1,                                          && \forall i \in V, \label{eq:time_window_select}\\
& t_{i} \geq \sum_{m \in W_i} y_{im} \cdot e_{im},                      && \forall i \in C, \label{eq:time_window_left} \\
& t_{i} \leq \sum_{m \in W_i} y_{im} \cdot l_{im},                      && \forall i \in C, \label{eq:time_window_right} \\
& x_{ijk}, y_{im} \in \{0, 1\}                                          && \forall i, j \in V, k \in K, m \in W_{i}, \label{eq:binary}
\end{align}

\begin{figure}
  \centering
  \vspace{-1.1em}
  \includegraphics[width=0.55\textwidth]{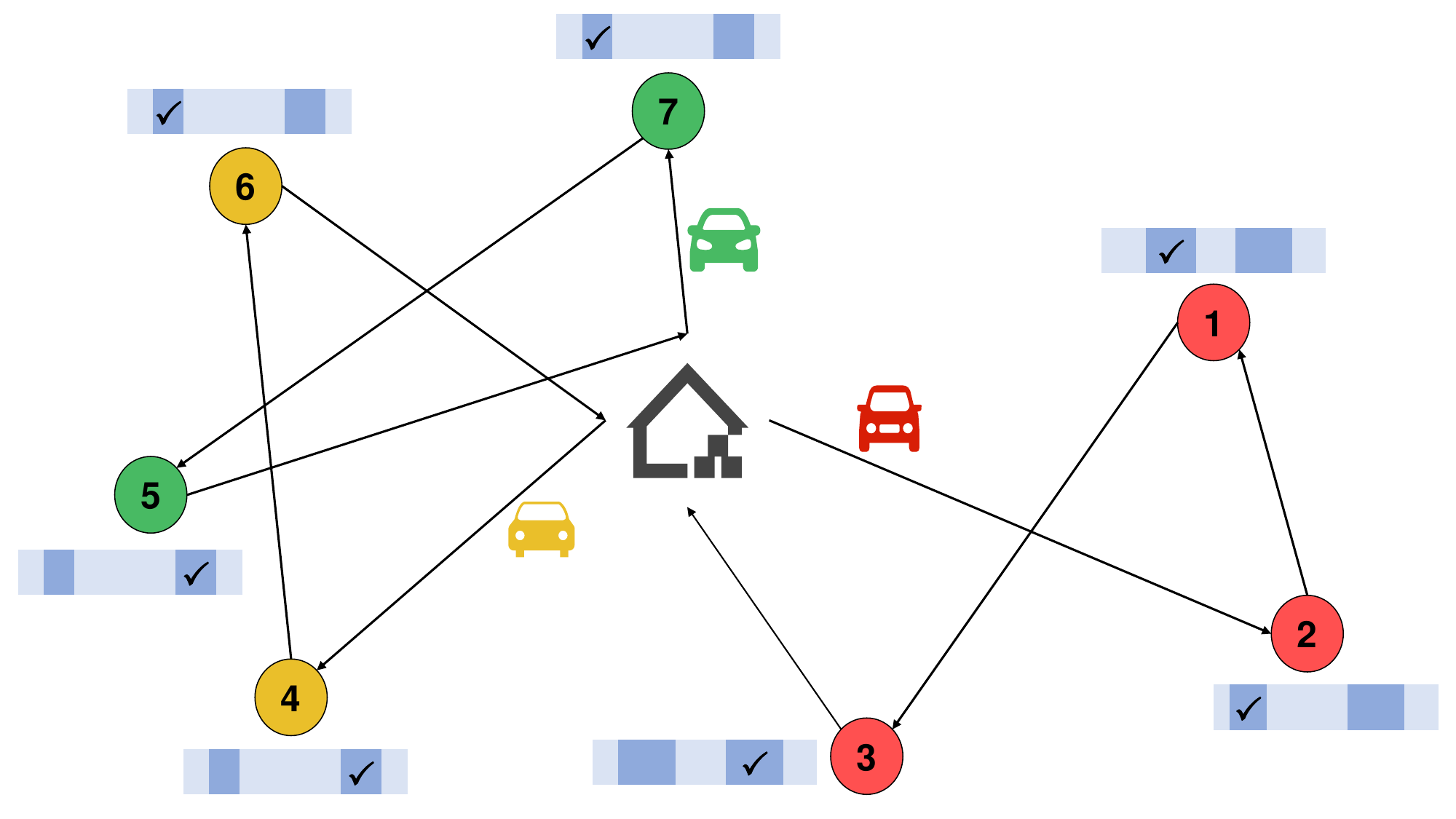}
  \caption{An illustration of VRPMTW, where each customer features two optional time windows.}
  \label{fig:illustration}
  \vspace{-1em}
\end{figure}


In particular, the objective function \eqref{eq:obj} aims to minimize the overall travel cost measured by the total distance traveled. Constraint \eqref{eq:customer_visit_once} ensures that each customer is visited exactly once by one vehicle. Constraints \eqref{eq:strat_at_depot} - \eqref{eq:end_to_depot} indicate that every vehicle must start its route from the depot and ultimately return to it. Constraint \eqref{eq:flow_conservation} preserves the continuity of vehicle flow, guaranteeing that, for each vehicle and every customer node, the inflow and outflow remain balanced. Constraints \eqref{eq:capacity} ensures that the total demand of customers visited by a vehicle does not exceed its capacity. Meanwhile, constraints \eqref{eq:subtour_elimination} represents the passage of time during vehicle travel and service, while simultaneously eliminating the subtours. Each vehicle‑specific time chain prevents disjoint subtours with the same vehicle. To deactivate this constraint when arc $(i,j)$ is not chosen, a sufficiently large constant $M$ is employed. Furthermore, constraint \eqref{eq:time_window_select} indicates that only one time window is selected for each customer. And constraints \eqref{eq:time_window_left}-\eqref{eq:time_window_right} guarantee that the service start time for each customer falls squarely within the chosen time window. Finally, constraint \eqref{eq:binary} restricts the range of values that the decision variables can take. For clarity, a toy example is provided in Figure \ref{fig:illustration}.

\section{Methodology}

\subsection{Initial solution}
To initiate the iterative process of search-based algorithms, a feasible initial solution is essential as it provides a crucial foundation for exploring and improving solutions in subsequent iterations. Algorithm~\ref{alg:greedy_vrptw} fulfills this requirement by generating a feasible initial solution for the VRPMTW instance. It implements a greedy algorithm that methodically constructs routes by iteratively selecting the nearest customer who satisfies both vehicle capacity and time window constraints. Beginning at the depot, routes are sequentially developed, ensuring feasibility with respect to both load and timing limitations. When no additional customers can be feasibly incorporated into the current route, a new route is initialized. While this greedy approach can not guarantee optimality, it is computationally efficient and establishes a solid foundation from which various neighborhood search algorithms can iteratively explore the solution space to identify superior alternatives.

\begin{algorithm}[h]
    \caption{Greedy Construction}
    \label{alg:greedy_vrptw}
    \setstretch{1.5}
    \begin{algorithmic}[1]
        \STATE \textbf{Input:} Customer set, depot location, vehicle capacity, time windows
        \STATE \textbf{Initialize:} Set current route $r \gets [0]$, load $q \gets 0$, time $t \gets 0$\\
        \quad\quad\quad Remaining customers $C \gets$ all customers, Routes $R \gets \emptyset$
        
        \WHILE{$C \neq \emptyset$}
            \STATE Find the nearest feasible customer $c \in C$ satisfying:\\
            \quad\quad Capacity: $q + d_c \leq Q$\\
            \quad\quad Time window: feasible arrival time for $c$
            \IF{$c$ is found}
                \STATE Append $c$ to $r$, update $q$, $t$, and current location
                \STATE Remove $c$ from $C$
            \ELSE
                \STATE Append $r + [0]$ to $R$
                \STATE Reset $r \gets [0]$, $q \gets 0$, $t \gets 0$
            \ENDIF
        \ENDWHILE

        \STATE \textbf{Output:} Routes $R$ covering all customers\\
    \end{algorithmic}
\end{algorithm}

\subsection{Variable neighborhood search}

Variable neighborhood search, introduced by \cite{hansen2001variable}, is a metaheuristic optimization framework designed to systematically exploit multiple neighborhood structures to avoid entrapment in local optima. In contrast to conventional local search algorithms that rely on a single neighborhood structure, VNS strategically switches among different neighborhood structures throughout the search process. The rationale behind VNS is grounded in three foundational observations: (1) A local optimum with respect to one neighborhood structure might not represent a local optimum under another; (2) A global optimum must necessarily be a local optimum with respect to all neighborhood structures; (3) Local optima corresponding to different neighborhood structures are frequently located in proximity within the solution space.

These insights highlight the advantages of methodically transitioning between neighborhood structures to enhance exploration capabilities and ultimately discover superior solutions. Our VNS implementation for VRPMTW follows the standard framework with three principal components: initialization, shaking, and local search. Initially, we establish a predefined set of neighborhood structures $\mathcal{N}_k$ $(k = 1, 2, ..., k_{\max})$, and generate an initial solution using the greedy construction in Algorithm~\ref{alg:greedy_vrptw}.

\begin{figure}
  \centering
  \includegraphics[width=0.48\textwidth]{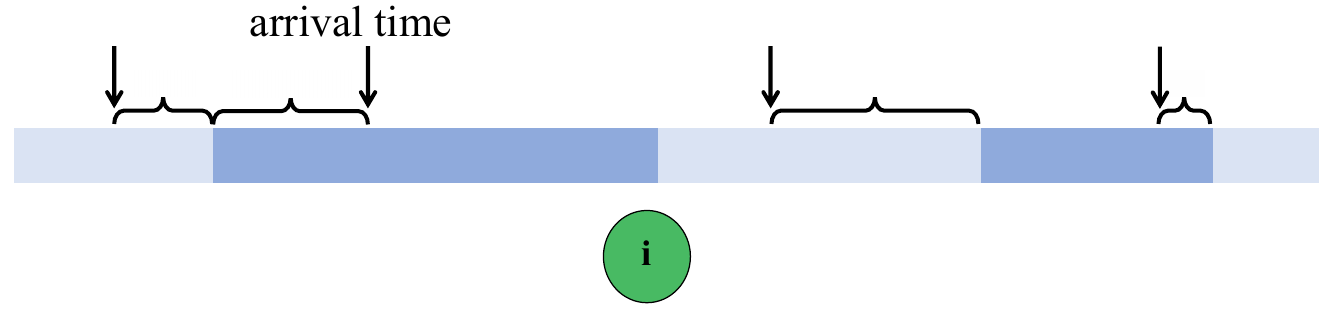}
  \caption{Illustration of fitness calculation for node i. The location of the arrow indicates the time when the vehicle arrived at the node $i$, and the length covered by the brace is the value of fitness.}
  \label{fig:fitness}
  \vspace{-1em}
\end{figure}

\paragraph{Shaking.}
To effectively navigate the solution space with multiple time window constraints, we introduce a fitness metric (Equation \eqref{eq:fitness}, illustrated in Figure \ref{fig:fitness}) that quantifies temporal flexibility. For each customer, this metric measures the deviation between the actual arrival time and the optimal service time within available time windows. Higher fitness values indicate greater temporal flexibility, while lower values suggest the solution might be near a local optimum with limited improvement potential. During the shaking phase, our operator selectively disrupts 20\% of customers with the highest fitness values, strategically reinserting them into alternative routes while preserving solution feasibility. This approach effectively balances exploration with the maintenance of high-quality solution characteristics.

\begin{equation}
\label{eq:fitness}
\text{fitness}(i) = 
\begin{cases}
\min(a_i - e_{im}, l_{im} - a_i) & \text{if }\exists m \in W_i \text{ s.t. } a_i \in [e_{im}, l_{im}] \\
\min_{m \in W_i} (e_{im} - a_i) \text{ s.t. } e_{im} > a_i & \text{otherwise}
\end{cases}
\end{equation}

\paragraph{Local search.}
Following the shaking phase, intensive local search examines the neighborhood of the perturbed solution to identify possible improvements. The overall performance and effectiveness of VNS critically depend on the choice of the operator set and their sequence, as these designs determine the depth and breadth of local search within the solution space. Our implementation employs a diverse set of neighborhood operators, including intra-route operators that reorganize node sequences within single routes, and inter-route operators that facilitate exchanges between different routes. Table \ref{tab:neighborhood_operators} provides a detailed enumeration of these strategically chosen operators. The operators are applied sequentially in a deterministic order, with the search advancing to the next neighborhood only when no further improvements can be found in the current one.

\begin{table}[h]
    \renewcommand{\arraystretch}{1.35}
    \centering
    \caption{Neighborhood operators for local search}
    \label{tab:neighborhood_operators}
    \resizebox{1.0\textwidth}{!}{
    \begin{tabular}{m{2cm}|m{3cm}|m{12cm}}
        \hline
        \textbf{Class} & \textbf{Name} & \textbf{Details} \\
        \hline
        \multirow{2}{*}{\parbox{2cm}{Intra-route}}
            & 2-Opt         & Remove two edges and reconnect their endpoints \\ \cline{2-3}
            & Move(1)   & Move a customer in the route to a new position \\
        \cline{1-3}
        \multirow{4}{*}{\parbox{2cm}{Inter-route}}
            & 2-Opt*      & Exchange the tails of two routes \\   \cline{2-3}
            & Swap(m) & Exchange segments of length $m\ (m = 1, 2, 3)$ between two routes \\ \cline{2-3}
            & Swap(m,n) &Exchange segments of length $m$ and $n\ (m,n = 1,2,3,m\neq n)$ between two routes. \\ \cline{2-3}
            & Relocate(m)   & Relocation of a segment of length $m\ (m = 1, 2, 3)$ from a route to another \\ 
        \hline
    \end{tabular}
    }
\end{table}

The pseudo-code for the VNS algorithm is detailed in Algorithm \ref{alg:basic_vns}.

\begin{algorithm}[h]
    \caption{Variable Neighborhood Search (VNS)}
    \label{alg:basic_vns}
    \setstretch{1.5}
    \begin{algorithmic}[1]
        \STATE \textbf{Input:} Sequence of neighborhood structures $\mathcal{N}=[\mathcal{N}_1, \mathcal{N}_2, \ldots, \mathcal{N}_{k_{max}}]$, initial solution $x_0$
        \STATE \textbf{Initialize:} Best solution $x \gets x_0$
        
        \REPEAT
            \STATE Set $k \gets 1$
            \WHILE{$k \leq k_{\max}$}
                \STATE \textit{Shaking:} Generate a random solution $x' \in \mathcal{N}_k(x)$
                \STATE \textit{Local search:} Apply local search method and obtain $x''=LS(\mathcal{N}_k, x')$
                \STATE \textit{Move or not:} 
                \IF{$f(x'') < f(x)$}
                    \STATE Set $x \gets x''$
                    \STATE Set $k \gets 1$ \COMMENT{Return to the first neighborhood}
                \ELSE
                    \STATE Set $k \gets k + 1$ \COMMENT{Move to the next neighborhood}
                \ENDIF
            \ENDWHILE
        \UNTIL{termination condition is met}
        
        \STATE \textbf{Output:} Best found solution $x$
    \end{algorithmic}
\end{algorithm}

\subsection{Adaptive variable neighborhood search}

The basic VNS often employs fixed neighborhood switching orders, which may lead to unnecessary exploration of less effective neighborhoods, thereby reducing overall performance. A natural extension to enhance its adaptability is to dynamically adjust the priority or selection likelihood of these operators during the search process. One well-established strategy for achieving this is a weight-based Adaptive VNS (AVNS) algorithm. This approach assigns a numerical weight to each operator, reflecting its perceived utility based on past performance within the current search. These weights are then used to determine the exploration sequence in subsequent iterations, aiming to favor those that have recently been more successful in discovering better solutions.

The AVNS maintains a weight vector $W=[w_1, w_2, \ldots, w_{k_{max}}]$, where each $w_k$ corresponds to the performance history of the i-th neighborhood operator. Initially, all weights are assigned to $1$. After each iteration, the weight of each operator is updated according to: 
\[
w_k = 
\begin{cases}
w_k + 5, & \text{if } f(x'') < f(x) \\
\max\{0, w_k - 1\}, & \text{if } f(x'') >= f(x)
\end{cases}
\]
where $x$ denotes the best solution identified thus far and $x''$ is the local optimum of $\mathcal{N}_k$. If the operator yields a new global best solution, the weight is increased by 5 units. Conversely, it imposes a penalty of 1 units on operators that fail to produce any improvement. This adaptive mechanism increases the weights of operators that successfully generate improved solutions, thereby enhancing their priority in subsequent iterations. Operators that consistently fail to yield improvements experience a gradual reduction in their weights.

This weight-based adaptive mechanism enables the VNS algorithm to learn from past experiences and progressively focus on more effective neighborhood structures for the specific problem instance being solved. However, this approach still has limitations as it primarily relies on historical performance without considering the the search progression stage—where improvements naturally become scarcer in later iterations, leading to potentially unfair penalization of operators that might still be valuable despite temporarily failing to yield improvements. This limitation motivates our proposed RL-based approach, which can dynamically adapt the selection of neighborhood structures based on the current solution state and learned patterns, potentially leading to more intelligent and effective exploration of the solution space.

\subsection{Reinforcement learning-based AVNS}
RL functions through an iterative cycle where an agent observes state $s_t$, selects action $a_t$ based on policy $\pi(a|s)$, and receives reward $r_t$ as the environment transitions to state $s_{t+1}$. The agent, implemented as a neural network, updates its parameters $\theta$ using experience tuples $(s_t, a_t, r_t, s_{t+1})$. Through gradient-based optimization, the agent progressively improves its decision-making to maximize expected cumulative rewards.


\begin{figure}
  \centering
  \includegraphics[width=1.0\linewidth]{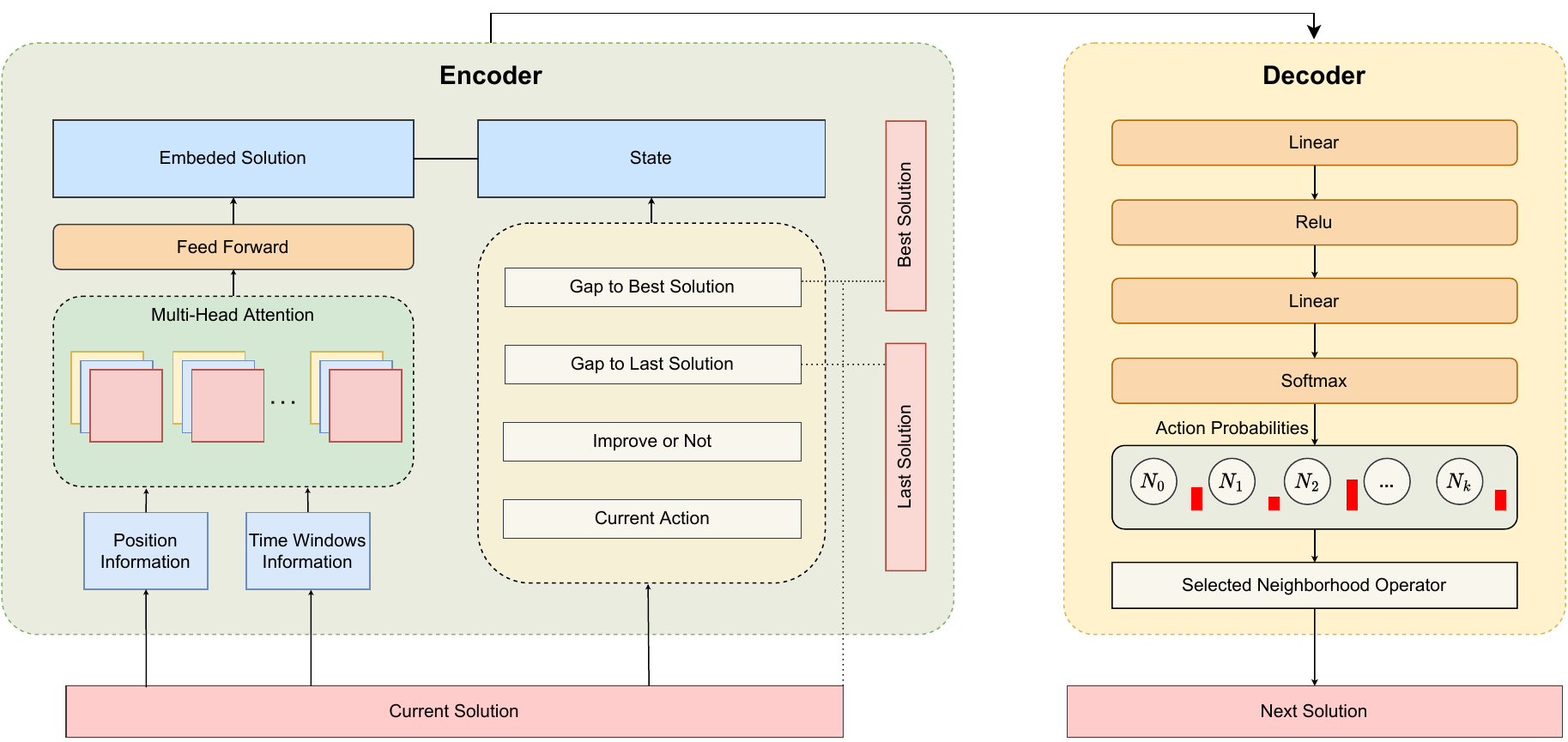}
  \caption{The architechture of our policy network in RL-AVNS.}
  \label{fig:network}
\end{figure}

\paragraph{States.}
The state representation in our model consists of two main components: information of the current solution and its relation to historical performance. The first component encompasses both static data (customer coordinates, time window constraints) and dynamic data (arrival times, sequence of predecessor and successor nodes). The second component evaluates the current solution by comparing it with both the previous solution and the best solution obtained so far, including an indicator signifying whether the current solution represents an improvement.

\paragraph{Actions.}
The action space comprises the set of neighborhood operators defined in Table \ref{tab:neighborhood_operators}, with each action corresponding to a specific operator. During each iteration, the agent selects an operator from this discrete action space, and the associated local search procedure is executed until a superior solution is identified. 

\paragraph{Policy network.}
The policy network in our RL-AVNS framework, depicted in Figure \ref{fig:network}, employs a transformer-based architecture with multi-head attention mechanisms \citep{vaswani2017attention} to encode state information. The encoder processes input features such as node positions and time window constraints through self-attention layers, capturing contextual dependencies within the problem instance. Subsequently, the decoder employs a feed-forward neural network generates raw scores for all neighborhood operators. These scores are normalized via a softmax layer to produce a probability distribution over the action space, enabling stochastic selection of operators according to their likelihood ratios.

\paragraph{Rewards.}
The reward mechanism is mathematically formulated as:
\[
r = [f(x_t) - f(x_{t+1})] - 100\cdot t_{run},
\]
where $t_{\text{run}}$ denotes the computational runtime. And then we bound the reward within the interval $[-10,10]$ to prevent reward magnitude explosion that could destabilize the learning process. It is worth mentioning that the runtime penalty term introduces the computational efficiency as an explicit optimization criterion, encouraging the algorithm to discover not only high-quality but also computationally tractable solution pathways. The coefficient 100 properly calibrates the trade-off between solution quality and computational economy, ensuring that marginal improvements requiring excessive computational resources are appropriately discouraged. This balanced reward formulation enables the reinforcement learning framework to develop a search policy that effectively navigates the solution space while maintaining algorithmic efficiency, which is a critical consideration for large-scale combinatorial optimization problems where computational resources are constrained.


\paragraph{Training.}
We train our RL-AVNS algorithm using Proximal Policy Optimization (PPO) \citep{schulman2017proximal}, a state-of-the-art policy gradient method known for its stability and effectiveness. During each training episode, the current policy network $\pi_\theta$ dynamically determines the execution order of neighborhood operators based on the given state representation $s_t$. The executed actions produce a trajectory comprised of state-action-reward sequences which is denoted as $\mathcal{D}$. The discounted returns $G_t$ are calculated as:
\[
G_t = \sum_{l=0}^{T-t-1} \gamma^l r_{t+l},
\]
where $\gamma \in (0,1]$ is the discount factor and $T=|\mathcal{D}|$ is the length of the trajectory. We use Generalized Advantage Estimation (GAE) \citep{schulman2015high} to compute advantage estimates $A_t$ as follows:
\[
A_t = \sum_{l=0}^{T-t-1} (\gamma\lambda)^l \delta_{t+l},\quad\text{with}\quad \delta_{t} = r_t + \gamma V_{\phi}(s_{t+1}) - V_{\phi}(s_t),
\]
where $\lambda \in [0,1]$ is the GAE smoothing parameter, $V_{\phi}$ is the value function parameterized by $\phi$, and $\delta_t$ is the temporal-difference residual error.

The policy network $\pi_\theta$ is updated by maximizing the PPO clipped surrogate objective function:
\[
J_{\text{PPO}}(\theta) = \frac{1}{T}\sum_{(s_t,a_t)\in\mathcal{D}} \min\left(
\frac{\pi_\theta(a_t|s_t)}{\pi_{\theta_{\text{old}}}(a_t|s_t)}A_t,\;
\text{clip}\left(\frac{\pi_\theta(a_t|s_t)}{\pi_{\theta_{\text{old}}}(a_t|s_t)}, 1-\epsilon, 1+\epsilon\right)A_t
\right),
\]
where $\epsilon$ is a hyperparameter controlling the clipping range, $\pi_{\theta_{\text{old}}}$ denotes the policy parameters before the current update, and $\mathcal{D}$ represents the collected trajectory dataset. Simultaneously, the value network $V_{\phi}(s_t)$ is trained by minimizing the mean squared error loss between predicted values and computed discounted returns:
\[
L_V(\phi) = \frac{1}{T}\sum_{(s_t,G_t)\in\mathcal{D}}\left(V_{\phi}(s_t)-G_t\right)^2.
\]

The details of the training process are presented in Algorithm \ref{alg:PPO-train} in Appendix. Through iterative application of these updates, our training procedure systematically enhances the policy network's ability to adaptively select the neighborhood operator, achieving a desirable trade-off between solution quality and computational efficiency in solving VRPMTW.

\begin{algorithm}[h]
    \caption{Reinforcement Learning-based Adaptive Variable Neighborhood Search (RL-AVNS)}
    \label{alg:RL-AVNS}
    \setstretch{1.5}
    \begin{algorithmic}[1]
        \STATE \textbf{Input:} A set of neighborhood operators for shaking or local search; \\
    \quad\quad\quad Initial solution $x_0$, maximum iterations $T$, trained policy network $\pi$
        
        \STATE \textbf{Initialize:} Set $t \gets 0$, best solution $x \gets x_0$, initial state $s_0$
        
        \WHILE{$t < T$}
            \STATE Evaluate policy network $\pi$ with input state $s_t$ to obtain the probability for each operator
            \STATE Sample an action $a_t$ and determine the local search operator $\mathcal{N}_{a_t}$
            \STATE $x \gets \text{LS}(\mathcal{N}_{a_t}, x)$ \COMMENT{Execute the local search process}
            \STATE Calculate next state $s_{t+1}$
            \STATE Set $t \gets t + 1$, $s_t \gets s_{t+1}$ 
        \ENDWHILE
        
        \RETURN Best solution $x$
    \end{algorithmic}
\end{algorithm}

\paragraph{Execution.}
After obtaining a trained policy network through the training procedure, we integrate this policy within the execution phase of our RL-AVNS framework. Specifically, the trained model is utilized online to guide the adaptive selection of neighborhood operators during the solution search process, as outlined in Algorithm \ref{alg:RL-AVNS}.

Starting from an initial feasible solution $x_0$, at each iteration $t$, the algorithm captures the current situation through state representation $s_t$, encoding both solution quality indicators and structural problem information. Then, the trained policy network $\pi$ processes $s_t$ to generate probability distributions over all neighborhood operators. Based on the probability, it samples an action $a_t$ that determines which neighborhood operator $\mathcal{N}_{a_t}$ to apply.

With this selected operator, we perform local search to improve upon the current best solution. After the local search concludes, the resulting solution's characteristics and the quality improvement obtained are leveraged to update the state to $s_{t+1}$. The iterative process repeats until reaching the predefined maximum iteration limit $T$, at which point the algorithm terminates and returns the best-found solution.

\section{Experiments}

\subsection{Experiment setup}

\paragraph{Data generation.}
In this study, we apply the VRPMTW to a novel and practical scenario: the replenishment of unmanned vending machines. This represents the first known application of VRPMTW in this context, as existing literature does not address the specific operational characteristics of vending machine replenishment. Unlike conventional problems, the time windows in our scenario follow specific distribution patterns rather than random or uniform distributions. This specificity arises because unmanned vending machines are typically located in venues with particular commercial functions, such as schools or fitness centers. Due to these venues' operational constraints, merchandise replenishment cannot be performed arbitrarily, resulting in time windows that cluster within several predetermined time periods.

Based on distribution data of a major unmanned vending machine manufacturer in the market, we have identified three primary replenishment windows: morning (6:00-9:00), midday (11:00-14:00), and evening (17:00-20:00). For each vending machine location, depending on its operational characteristics, two or more of these replenishment opportunities are available. For instance:
\begin{itemize}
    \item Vending machines in fitness centers can typically be replenished in the morning or midday, as customers usually concentrate their exercise activities in the evening hours.
    \item School-based vending machines generally accommodate all three replenishment windows, as these times avoid peak class hours.
    \item Machines in commercial districts often have morning and evening replenishment windows, when customer traffic is lower and daytime consumption has depleted inventory, making these periods ideal for restocking.
\end{itemize}
Based on these practical insights, our data generation process randomly assigns each customer node two or more time windows from the three identified periods, closely approximating real-world conditions. This approach ensures our problem instances reflect real-world operational constraints in vending machine replenishment logistics.

Specifically, customer locations in our experimental setup are uniformly distributed across a $[0, 100]$ range. Each customer's demand, $q_i$, is drawn from a truncated normal distribution, $\mathcal{N}(15, 10^2)$, with truncation interval $[1, 42]$. The vehicle capacity is fixed at $Q = 100$, and the overall time horizon is set from $[e_0, l_0] = [0, 1000]$, with a consistent service time of $s_i \equiv 10$. All time windows $[e_{im}, l_{im}]$ for each customer are generated sequentially from the earliest to the latest, following the distribution pattern identified from real-world vending machine operations.

\paragraph{Training and hyper-parameters.}
We utilized the Adam optimizer with a learning rate of 0.001 for training. The encoder integrates state and solution information, including features such as node position and time windows. This information is processed using multi-head attention blocks with a dimensionality of 128, followed by a feed-forward layer. The decoder consists of two hidden layers, each with 256 dimensions, interspersed with a Rectified Linear Units (ReLU) activation layer. A softmax funtion then computes the probability of operators, thereby sampling one (as illustrated in Figure \ref{fig:network}). The model underwent training over 2000 epochs, with each epoch consisting of 20,000 steps, during which the model predicts actions for either shaking or local search processes. The method was implemented using Python, and experiments were conducted on a computer equipped with a single NVIDIA GeForce RTX 4090 GPU and an Intel i9-14900 CPU. 

\paragraph{Evaluation metrics.}
To comprehensively evaluate the performance of different methods, we employ four key metrics, all computed as the average over 100 randomly generated instances. (1) Total Length (Length) measures the sum of travel distances for all routes. (2) Total Duration (Duration) quantifies the cumulative time required for all vehicles to complete their routes and return to the depot. (3) Number of Vehicles Used (K) counts the total number of vehicles employed, with each vehicle’s service concluded upon returning to the depot. (4) Solving Time ($t_{\text{solve}}$) records the execution time required for each method to solve all instances, reflecting computational efficiency. All metrics are reported as averages across the 100 test cases, enabling fair and statistically robust comparisons between methods.

\paragraph{Compared algorithms.} 
To robustly evaluate our method, we compare it against several established solvers across different configurations of time windows. For small-scale cases, we employ Gurobi to solve the mathematical formulation presented previously in Section~\ref{sec:model}. Additionally, we include Random VNS (RVNS), which employs a random policy for selecting neighborhood operators, as well as AVNS, a traditional adaptive VNS method that adjusts operator selection based on historical performance, to clearly demonstrate the effectiveness of our reinforcement learning-based adaptive mechanism.

For the single time window scenario, we further compare our method with state-of-the-art learning-based solvers, specifically JAMPR \citep{JAMPR} and AM \citep{kool2018attention}. JAMPR is particularly relevant as a baseline because it represents one of the most advanced neural constructive heuristics specifically tailored for VRPTW problems, incorporating sophisticated attention mechanisms and masking strategies to effectively handle time window constraints. Moreover, we compare against LKH3, a state-of-the-art solver for its exceptional performance in classical VRPs, and OR-Tools \citep{perron2019or}, a well-known meta-heuristic solver renowned for its advanced local search techniques that efficiently address routing problems. However, since LKH3 and OR-Tools do not support multiple time window constraints, their evaluation is limited to single time window scenarios. 
Through this comprehensive set of comparisons, we aim to rigorously assess the performance and generalizability of our proposed method across varying complexities and problem structures within vehicle routing scenarios.

\paragraph{Inference setup.}
We evaluate the performance of RL-AVNS across the VRPTW and three scenarios of the VRPMTW, varying the number of time windows per customer (two, three, and mixed configurations). For each instance, the trained policy network guides the selection of neighborhood operators through 2,000 iterations, each of which is executed based on the preceeding ones. To show the generalization ability of our method, we only train it on the instances with 50 nodes ($N=50$) and 3 time windows, and then applly it universally across all settings. For AM, we report results using both greedy decoding and sampling (100 samples). Performance metrics, such as total length and total time, are averaged over 100 random instances unless specified otherwise.

\subsection{Results on small-scale instances}
To illustrate the complexity of VRPMTW and demonstrate the infeasibility of directly applying exact optimization methods, we conducted experiments using Gurobi, a state-of-the-art commercial solver, to solve the mathematical formulation presented in Section~\ref{sec:model}. We select two small-scale scenarios, with instance sizes $N=10$ and $N=20$, and set a computational time limit of 3600 seconds per instance. Each scenario included multiple time windows, where customers have either 2 or 3 feasible time windows.

Table~\ref{tab:small_scale} summarizes the results obtained from Gurobi. At the smallest scale, Gurobi successfully solved all instances to optimality within the one-hour time frame, indicating that exact approaches may still be viable for very small problem instances. However, when increasing the number of customers to 20, Gurobi could not find any feasible solution within the given time limit. This sharp deterioration in performance clearly highlights the limitations of exact methods in addressing the combinatorial complexity of VRPMTW, even for moderately-sized instances.

\begin{table}[!ht]
    \centering
    \caption{Results small-scale VRPMTW instances.}
    \label{tab:small_scale}
    \resizebox{0.7\textwidth}{!}{
    \begin{tabular}{l|cccc|cccc}
    \toprule
    \multirow{2}{*}{\textbf{\quad\quad Method}} & \multicolumn{4}{c|}{\textbf{N=10}} & \multicolumn{4}{c}{\textbf{N=20}} \\
    \cmidrule(lr){2-5} \cmidrule(lr){6-9}
     & Length & Duration & K & $t_\text{solve}$ & Length & Duration & K & $t_\text{solve}$ \\
     \cmidrule(lr){1-9}
        Gurobi & 283 & 1087 & 1.7 & 6s & / & / & / & / \\ 
        VNS & 306 & 1293 & 1.8 & 2m &  698 & 2541 & 4.0 & 9.4m \\

    \bottomrule
    \end{tabular}
    }
\end{table}

\subsection{Results on large-scale instances}
Given the computational intractability of exact methods demonstrated in small-scale scenarios, we systematically evaluate heuristic approaches for real-world large-scale problems. For OR-Tools, the solving time is configured at 600 seconds for $N=50$ instances and 3,600 seconds for $N=100$. Similarly, for LKH3, we set the number of runs at 10, adhering to the same time limits as OR-Tools. 

\paragraph{VRPTW.}
Table \ref{tab:VRPTW} displays the results of various methods in single time window scenario for instance sizes of 50 and 100. RL-AVNS significantly outperforms the neural solvers AM (both greedy and sampling) and JAMPR in terms of total length and total time. Interestingly, while JAMPR shows better performance for larger size, it is less effective for smaller ones compared to AM. And RL-AVNS excels over OR-Tools in total length but remains slightly behind the specialized solver LKH3. The relatively lower efficiency of RL-VANS may be attributed to its reliance on heuristic operators coded in Python. In contrast, solvers like LKH3 and OR-Tools, implemented in C++, benefit from inherent optimizations, while both AM and JAMPR gain performance enhancements from GPU acceleration.
It is worth to mention that RL method demonstrates remarkable improvements over its predecessor AVNS. Specifically, RL-AVNS reduces solution length by 6-9\%, travel time by approximately 10\%, and required vehicles by around 8\% across both instance sizes. Most impressively, the RL approach accelerates computation by 3-5 times compared to standard AVNS. This substantial improvement validates our reinforcement learning strategy for adaptive neighborhood selection, which effectively guides the search process while avoiding wasteful exploration.

\begin{table}[!ht]
    \centering
    \caption{Results for VRPTW. The line named "Gain" represents the improvement of RL-AVNS compared to AVNS.}
    \label{tab:VRPTW}
    \resizebox{0.99\textwidth}{!}{
    \begin{tabular}{l|cccc|cccc}
    \toprule
    \multirow{2}{*}{\textbf{\quad\quad Method}} & \multicolumn{4}{c|}{\textbf{N=50}} & \multicolumn{4}{c}{\textbf{N=100}} \\
    \cmidrule(lr){2-5} \cmidrule(lr){6-9}
     & Length & Duration & K & $t_\text{solve}$ & Length & Duration & K & $t_\text{solve}$ \\
     \cmidrule(lr){1-9}
        OR-Tools \citep{perron2019or}       & 1804 & 8914 & 12.6 & 10m & 2761 & 13864 & 18.9 & 1h  \\ 
        LKH3 \citep{helsgaun2017extension}  & 1417 & 7518 & 11.3 & 9m & 2473 & 12969 & 19.7 & 1h \\ \cmidrule(lr){1-9}
        AM (greedy) \citep{kool2018attention}  & 2031 & 11518 & 15.5 & 3s & 3389 & 17495 & 23.1 & 5s \\
        AM (sampling) \citep{kool2018attention}  & 1970 & 11034 & 14.8 & 2m & 3303 & 16362 & 21.4 & 4m \\
        JAMPR \citep{JAMPR}     & 2241 & 14861 & 21.1 & 3s & 2679 & 19406 & 24.8 & 5s \\ \cmidrule(lr){1-9}
        VNS & 1873 & 10921 & 15.0 & 31m & 3146 & 19974 & 26.3 & 2.0h \\
        RVNS & 1981 & 11862 & 15.3 & 48m & 3394 & 21906 & 27.9 & 2.4h \\
        AVNS & 1692 & 10931 & 13.8 & 1.3h & 2718 & 16738 & 22.0 & 4.1h \\
        RL-AVNS & 1538 & 9786 & 12.6 & 15m & 2548 & 15032 & 20.2 & 1.1h \\
        \cmidrule(lr){1-9}
        Gain & 9.1\% &10.5\% & 8.7\% & 80.8\% & 6.3\% & 10.2\% & 8.2\% & 73.2\% \\
    \bottomrule
    \end{tabular}
    }
\end{table}

\paragraph{VRPMTW.}
We consider three scenarios: instances with 2 time windows (2-TW), instances with 3 time windows (3-TW), and instances with mixed time windows (Mix-TW), where customers may have either 2 or 3 time windows. As indicated in Table \ref{tab:VRPMTW}, RL-AVNS consistently outperforms all baseline methods across different problem settings. For the 2-TW scenario, RL-AVNS achieves length reductions of 3.3\% and 10.8\% compared to AVNS for $N=50$ and $N=100$, respectively. In the 3-TW scenario, improvements reach 14.5\% and 12.3\%, while the mixed time window setting shows gains of 6.9\% and 5.9\%.

Most notably, RL-AVNS dramatically reduces computational time by 65\% to 77\% compared to AVNS across all configurations. This efficiency gain stems from AVNS's limitation of selecting operators solely based on solution quality without considering execution time. Consequently, AVNS favors operators that might yield better solutions but require significantly longer computation, particularly problematic in complex multiple time window scenarios. In contrast, RL-AVNS learns to balance solution quality with computational efficiency. The results demonstrate that as problem complexity increases, the advantages of reinforcement learning for guiding the search process become increasingly pronounced. The learned policy effectively navigates the solution space by making intelligent decisions about neighborhood operator selection, thus achieving a more favorable trade-off between solution quality and computational efficiency.

\begin{table}[!ht]
    \centering
    \vspace{-1em}
    \caption{Results for VRPMTW. The line named "Gain" represents the improvement of RL-AVNS compared to AVNS.}
    \label{tab:VRPMTW}
    \resizebox{0.85\textwidth}{!}{
    \begin{tabular}{ll|cccc|cccc}
        \toprule
        \multirow{2}{*}{} & \multirow{2.5}{*}{\textbf{Method}} & \multicolumn{4}{c|}{\textbf{N=50}} & \multicolumn{4}{c}{\textbf{N=100}} \\
        \cmidrule(lr){3-6} \cmidrule(lr){7-10}
        & & Length & Duration & K & $t_\text{solve}$ & Length & Duration & K & $t_\text{solve}$\\
        \cmidrule(lr){1-10}
        \multirow{5}{*}{\textbf{2-TW}}  
            ~ & VNS     & 1063  & 2792  & 5.9   & 51m  & 2086   & 4953  & 9.6   & 2.5h \\
            ~ & RVNS    & 1078  & 2825  & 6.0   & 56m  & 2127   & 5032  & 9.7   & 2.4h \\
            ~ & AVNS    & 1032  & 2731  & 5.6   & 1.5h & 1978   & 4827  & 9.3   & 6.8h \\
            ~ & RL-AVNS & 998   & 2597  & 5.3   & 21m   & 1763   & 4459  & 8.9   & 1.6h  \\
            ~ & Gain    & 3.3\% & 4.9\% & 5.4\% & 76.7\%  & 10.8\% & 7.6\% & 4.3\% & 76.5\% \\  
        \cmidrule(lr){1-10}
        \multirow{5}{*}{\textbf{3-TW}}  
            ~ & VNS     & 1157 & 2472   & 4.8   & 52m   & 1692  & 5247  & 8.7   & 2.3h \\ 
            ~ & RVNS    & 1192 & 2587   & 5.0   & 59m   & 1762  & 5401  & 9.1   & 2.6h \\ 
            ~ & AVNS    & 1086 & 2431   & 4.6   & 1.8h  & 1637  & 5093  & 8.3   & 5.9h \\ 
            ~ & RL-AVNS & 929   & 2307  & 4.3   & 36m   & 1435  & 4132  & 7.4   & 1.9h \\ 
            ~ & Gain    & 14.5\% & 5.1\% & 6.5\% & 66.7\%   & 12.3\% & 18.9\% & 10.8\% & 67.8\% \\
        \cmidrule(lr){1-10}
        \multirow{5}{*}{\textbf{Mix-TW}}  
            ~ & VNS     & 1142  & 2864  & 6.2   & 1.1h & 2215   & 5127  & 10.1  & 3.1h \\
            ~ & RVNS    & 1173  & 2953  & 6.4   & 1.3h & 2293   & 5216  & 10.3  & 3.4h \\
            ~ & AVNS    & 1097  & 2819  & 5.9   & 2.2h & 1979   & 4982  & 9.8   & 7.5h \\
            ~ & RL-AVNS & 1021  & 2654  & 5.5   & 39m  & 1862   & 4523  & 9.1   & 2.6h \\
            ~ & Gain    & 6.9\% & 5.9\% & 6.8\% & 70.5\%  & 5.9\% & 9.2\% & 7.1\% & 65.3\% \\
        \bottomrule
    \end{tabular}
    }
\end{table}

\begin{figure}[!ht]
    \centering
    \includegraphics[width=1.0\linewidth]{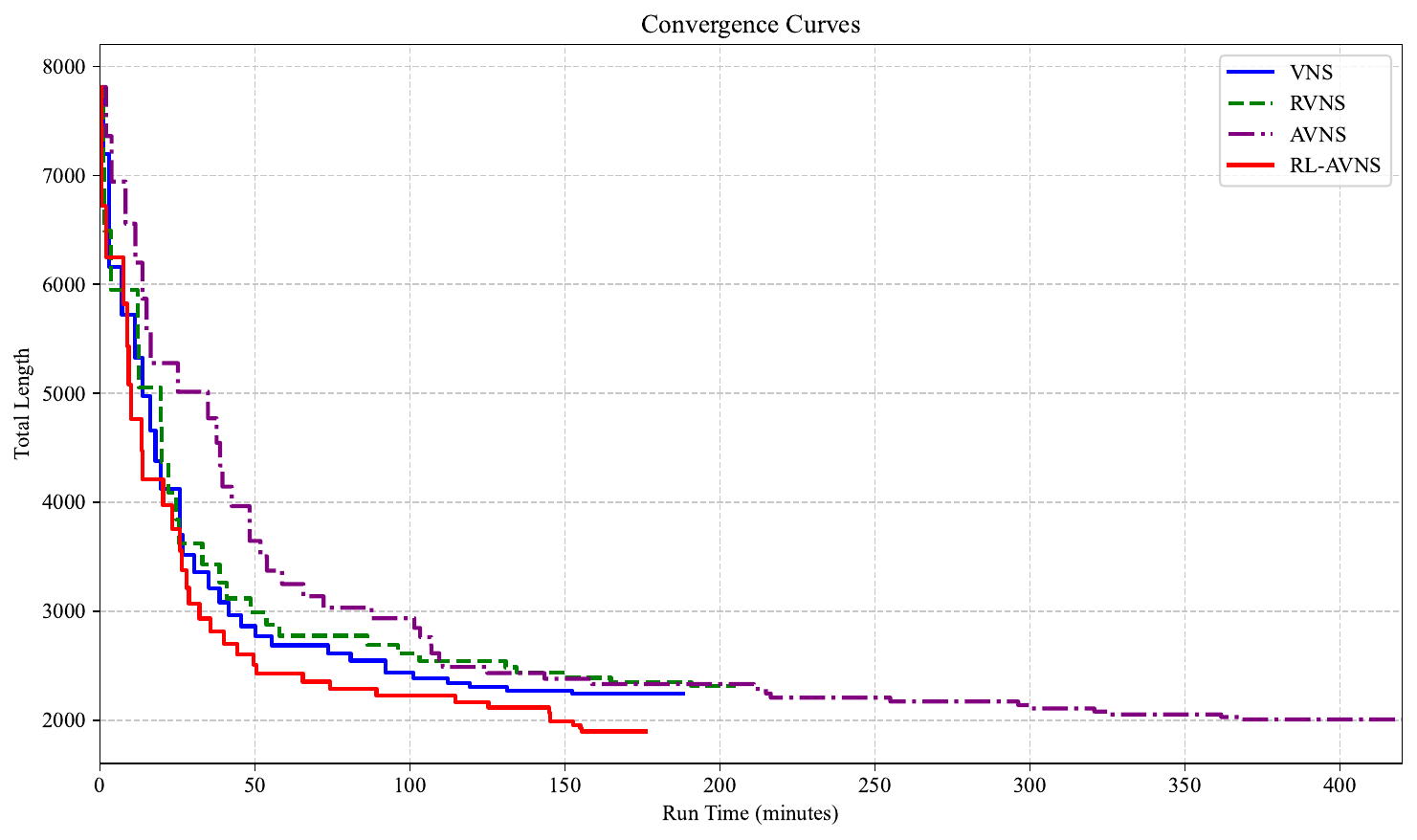}
    \vspace{-0.5em}
    \caption{Convergence curves for different algorithms on a Mix-TW instance with $N=100$ nodes.}
    \label{fig:convergence}
    \vspace{-0.5em}
\end{figure}

\paragraph{Convergence Process.}
Figure~\ref{fig:convergence} illustrates the convergence behavior of the four algorithms on a Mix-TW instance with $N=100$ nodes, plotting the total route length against run time in minutes. The graph clearly demonstrates the superior performance of RL-AVNS in terms of both convergence speed and solution quality. While all algorithms start from the same initial solutions by greedy algorithm, RL-AVNS exhibits a remarkably steeper descent curve, reaching a high-quality solution within the first 50 minutes. Particularly notable is AVNS's relatively poor early-stage performance, which supports our earlier observation about its tendency to select time-consuming operators without considering computational efficiency.  VNS and RVNS demonstrate better initial convergence than AVNS, but their improvement rates plateau earlier and reach inferior final solutions. In contrast, RL-AVNS maintains consistent improvement throughout the search process, highlighting its ability to make more effective operator selections that balance quality improvements with computational efficiency.

\begin{figure}[!ht]
    \centering
    \includegraphics[width=1.0\linewidth]{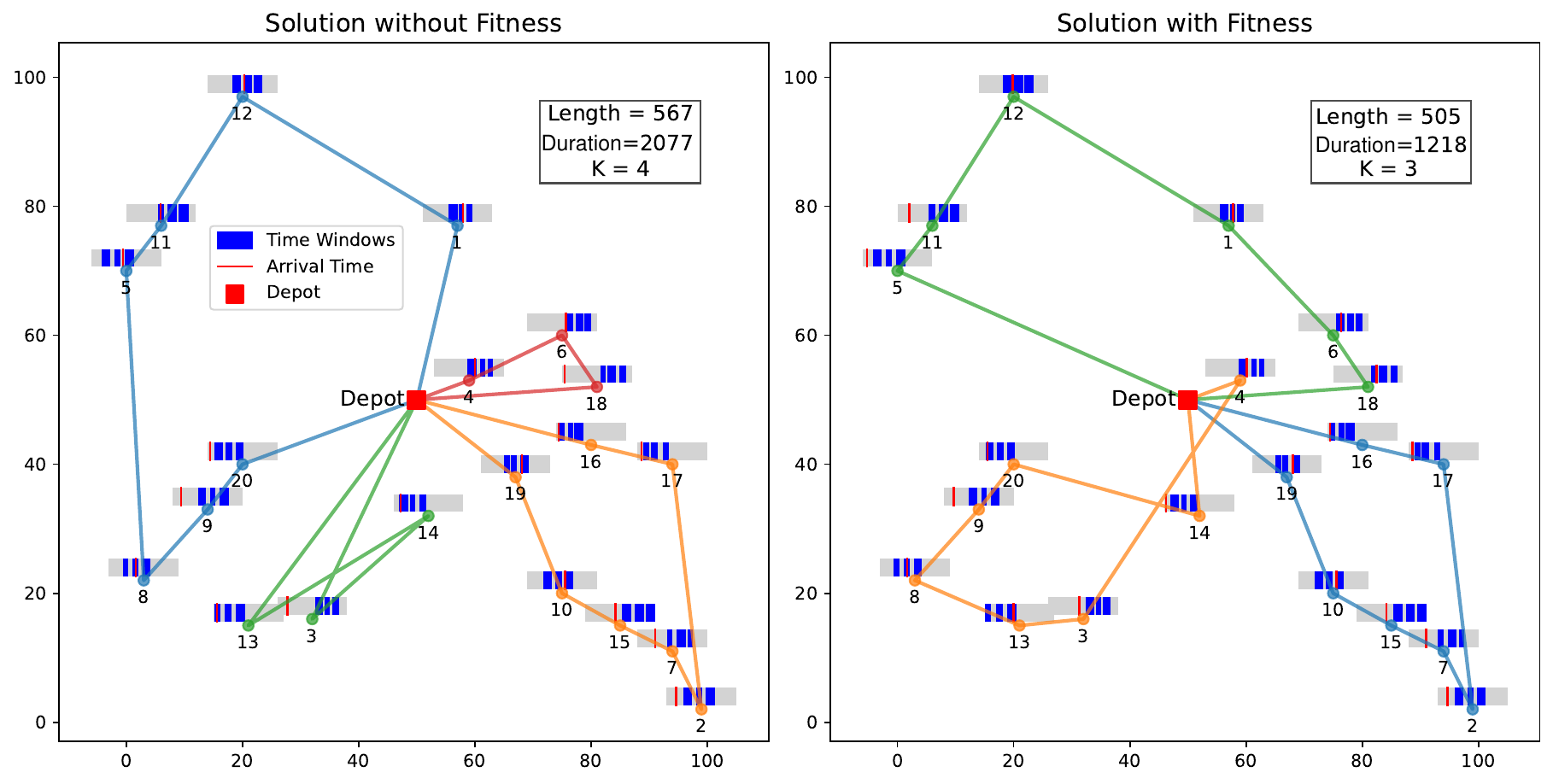}
    \vspace{-0.5em}
    \caption{The effect of fitness metric on the final solution.}
    \label{fig:solution}
    \vspace{-0.5em}
\end{figure}

\paragraph{Role of fitness.}
As depicted in Figure \ref{fig:solution}, incorporating the fitness metric into the shaking process significantly enhances solution quality. The figure presents two routing solutions for the same instance: one without the fitness metric (left) and one with the fitness-based approach (right). In the left subgraph, the red path shows prolonged waiting times after the vehicle's arrival, a result of late time windows for customers $(18, 6, 4)$. Conversely, the path in the right subgraph aligns more effectively with the customers’ time windows, demonstrating the practical benefits of applying the fitness metric to optimize the route. This reduction in both distance and duration demonstrates how the fitness-based shaking operator produces more efficient routing schedules that better synchronize with customer availability windows.

\section{Conclusion}
In this paper, we introduced RL-AVNS, a reinforcement learning-based adaptive variable neighborhood search method tailored to solve the VRPMTW, specifically applied to the replenishment logistics of unmanned vending machines. By dynamically adapting neighborhood operator selection using a trained policy network, RL-AVNS achieves significant improvements in both solution quality and computational efficiency compared to traditional methods. Additionally, the proposed fitness metric further enhances the shaking phase by intelligently identifying customers with greater temporal flexibility, facilitating more effective exploration of the solution space.

Comprehensive experimentation reveals that RL-AVNS consistently surpasses traditional metaheuristics and learning-based solvers. The results confirm that RL agent's adaptive decision-making capability significantly boosts the performance of neighborhood search heuristics in complex routing scenarios, highlighting its potential for broader applications in combinatorial optimization problems. Future improvements could focus on the generalization capability of RL-AVNS across various VRP variants beyond those with multiple time windows. Through this expansion of scope and further refinement of its computational efficiency, RL-AVNS would become an even more robust and versatile tool for solving complex routing problems in real-world logistics and transportation scenarios.

\bibliographystyle{cas-model2-names}

\bibliography{cas-refs}

\newpage

\appendix
\section{My Appendix}
\begin{algorithm}[htbp]
    \caption{Training Procedure for RL-AVNS}
    \label{alg:PPO-train}
    \setstretch{1.5}
    \begin{algorithmic}[1]
        \STATE \textbf{Input:} Learning rates for policy network $\alpha_\theta$ and value network $\alpha_\phi$, clipping parameter $\epsilon$, epochs per update $K$, number of episodes $E$, maximum steps per episode $T$
        \STATE \textbf{Initialize:} Policy network $\pi_\theta$, value network $V_\phi$, set $\theta_{\text{old}} \gets \theta$

        \FOR{$\text{episode} = 1,2,\dots,E$}
            \STATE Generate a new VRPMTW instance $P$
            \STATE Initialize solution $x_0$ and state $s_0$, set $t \gets 0$
            \STATE Initialize empty trajectories buffer $\mathcal{D} \gets \emptyset$
            
            \WHILE{$t < T$}
                \STATE Evaluate policy network $\pi_\theta$ given state $s_t$ to obtain the probability distribution 
                \STATE Sample an action $a_t$ and determine the local search operator $\mathcal{N}_{a_t}$
                \STATE $x_{t+1} \gets \text{LS}(\mathcal{N}_{a_t}, x_t)$
                \STATE Compute reward $r_t = [f(x_t) - f(x_{t+1})] - 100\cdot t_{\text{run}}$, clip $r_t$ to $[-10,10]$
                \STATE Observe new state $s_{t+1}$, store transition $(s_t, a_t, r_t, s_{t+1})$ in buffer $\mathcal{D}$
                \STATE Set $t \gets t+1$, $s_t \gets s_{t+1}$, $x_t \gets x_{t+1}$
            \ENDWHILE

            \STATE Compute discounted returns $G_t=\sum_{l=0}^{T-t-1} \gamma^l r_{t+l}$ 
            \STATE Compute advantage estimates $A_t = \sum_{l=0}^{T-t-1} (\gamma\lambda)^l \delta_{t+l},\quad\text{with}\quad \delta_{t} = r_t + \gamma V_{\phi}(s_{t+1}) - V_{\phi}(s_t),$

            \FOR{$k=1,2,\dots,K$}
                \STATE Update policy parameters by maximizing PPO objective:
                \[
                \theta \gets \theta + \alpha_\theta \nabla_\theta \frac{1}{T}\sum_{(s_t,a_t)\in\mathcal{D}} \min\left(\frac{\pi_\theta(a_t|s_t)}{\pi_{\theta_{\text{old}}}(a_t|s_t)}A_t, \text{clip}\left(\frac{\pi_\theta(a_t|s_t)}{\pi_{\theta_{\text{old}}}(a_t|s_t)},1-\epsilon,1+\epsilon\right)A_t\right)
                \]

                \STATE Update value function parameters by minimizing mean squared error:
                \[
                \phi \gets \phi - \alpha_\phi \nabla_\phi \frac{1}{T}\sum_{(s_t,G_t)\in\mathcal{D}}(V_\phi(s_t)-G_t)^2
                \]
            \ENDFOR
            \STATE Set $\theta_{\text{old}} \gets \theta$
        \ENDFOR

        \RETURN Optimized policy network parameters $\theta$
    \end{algorithmic}
\end{algorithm}

\printcredits

\end{document}